\documentclass[runningheads]{llncs}
\usepackage{graphicx}
\usepackage{subcaption}
\usepackage{amsmath}
\usepackage{amssymb}
\usepackage{booktabs}
\usepackage{enumitem}
\usepackage{multicol}
\usepackage{breakcites}
\usepackage{multirow}
\usepackage[T1]{fontenc}
\usepackage{wrapfig}
\usepackage{colortbl}
\usepackage{pifont}
\usepackage{soul}
\usepackage{hyperref}
\hypersetup{breaklinks,colorlinks}

\usepackage{booktabs}
\DeclareUnicodeCharacter{0300}{`}
\newcommand{\cmark}{\ding{51}}%
\newcommand{\xmark}{\ding{55}}

\newcommand{\inputlinedecomp}[1]{\includegraphics[width=0.14\linewidth]{figures/decomposition/all_images/input_#1.png} & \includegraphics[width=0.14\linewidth]{figures/decomposition/all_images/gt_#1.png} & \includegraphics[width=0.14\linewidth]{figures/decomposition/all_images/spair_#1.png} & \includegraphics[width=0.14\linewidth]{figures/decomposition/all_images/gmair_#1.png} & \includegraphics[width=0.14\linewidth]{figures/decomposition/all_images/space_#1.png} & \includegraphics[width=0.14\linewidth]{figures/decomposition/all_images/ast_#1.png} & \includegraphics[width=0.14\linewidth]{figures/decomposition/all_images/dti_#1.png}\\}
\newcommand{\changedettable}[1]{\includegraphics[width=0.14\linewidth]{figures/change_detection/congealing/#1/input_normal_0.jpg} & \includegraphics[width=0.14\linewidth]{figures/change_detection/congealing/#1/input_changed_0.jpg} & \includegraphics[width=0.14\linewidth]{figures/change_detection/congealing/#1/ground_truth_changed_0.jpg} & \includegraphics[width=0.14\linewidth]{figures/change_detection/naive/#1/diff_image_changed_0.jpg} & \includegraphics[width=0.14\linewidth]{figures/change_detection/vae/#1/diff_image_changed_0.jpg} & \includegraphics[width=0.14\linewidth]{figures/change_detection/stae/#1/diff_image_changed_0.jpg} & \includegraphics[width=0.14\linewidth]{figures/change_detection/congealing/#1/diff_image_changed_0.jpg} }
\newcommand{\changedettablerot}[1]{\rotatebox[origin=c]{90}{\includegraphics[height=0.14\linewidth]{figures/change_detection/congealing/#1/input_normal_0.jpg}} & \rotatebox[origin=c]{90}{\includegraphics[height=0.14\linewidth]{figures/change_detection/congealing/#1/input_changed_0.jpg}} & \rotatebox[origin=c]{90}{\includegraphics[height=0.14\linewidth]{figures/change_detection/congealing/#1/ground_truth_changed_0.jpg}} & \rotatebox[origin=c]{90}{\includegraphics[height=0.14\linewidth]{figures/change_detection/naive/#1/diff_image_changed_0.jpg}} & \rotatebox[origin=c]{90}{\includegraphics[height=0.14\linewidth]{figures/change_detection/vae/#1/diff_image_changed_0.jpg}} & \rotatebox[origin=c]{90}{\includegraphics[height=0.14\linewidth]{figures/change_detection/stae/#1/diff_image_changed_0.jpg}} & \rotatebox[origin=c]{90}{\includegraphics[height=0.14\linewidth]{figures/change_detection/congealing/#1/diff_image_changed_0.jpg}} }
\NewDocumentCommand{\rot}{O{35} O{4em} m}{\makebox[#2][l]{\rotatebox{#1}{#3}}}%

\renewcommand{\arraystretch}{1.5}
\newcommand{\blockcomment}[1]{}
\definecolor{lightgray}{gray}{0.6}
\definecolor{lightergray}{gray}{0.9}
\def\gc{\color{lightgray}}

\newcolumntype{C}[1]{>{\centering\arraybackslash}p{#1}}

\begin{document}

\title{Historical Printed Ornaments: Dataset and Tasks}
\author{Sayan Kumar Chaki$^{*}$\inst{1}\orcidID{0000-0003-1390-1329} \and
Zeynep Sonat Baltaci$^{*}$\inst{2}\orcidID{0000-0001-6749-2407} \and
Elliot Vincent\inst{2,3}\orcidID{0009-0001-1713-2590
} \and Remi Emonet\inst{1,4}\orcidID{0000-0002-1870-1329} \and Fabienne Vial-Bonacci\inst{5}\orcidID{0000-0001-6202-6407} \and Christelle Bahier-Porte\inst{5}\orcidID{0000-0003-1775-0142} \and Mathieu Aubry\inst{2}\orcidID{0000-0002-3804-0193} \and Thierry Fournel\inst{1}\orcidID{0000-0002-1613-4594}}
\authorrunning{Chaki et al.}

\institute{Univ. Lyon, UJM-St-Etienne, CNRS, Institut d'Optique Graduate School, Inria, Laboratoire Hubert Curien UMR 5516, F-42023, Saint-Etienne, France \and
LIGM, Ecole des Ponts, Univ Gustave Eiffel, CNRS, Marne-la-Vallée, France \and
Inria, Ecole normale supérieure, CNRS, PSL Research University, Paris, France \and
IUF University Institute of France \and
UJM-St-Etienne, CNRS, IHRIM 5317, Saint-Etienne, France
\email{\{sayan.kumar.chaki,remi.emonet,fabienne.vial,christelle.porte, fournel\}@univ-st-etienne.fr, \{sonat.baltaci, elliot.vincent,mathieu.aubry\}@enpc.fr}}
\maketitle              
%
\begin{abstract}
This paper aims to develop the study of historical printed ornaments with modern unsupervised computer vision. We highlight three complex tasks that are of critical interest to book historians: clustering, element discovery, and unsupervised change localization. For each of these tasks, we introduce an evaluation benchmark, and we adapt and evaluate state-of-the-art models. Our \textit{Rey's Ornaments dataset} is designed to be a representative example of a set of ornaments historians would be interested in. It focuses on an XVIIIth century bookseller, Marc-Michel Rey, providing a consistent set of ornaments with a wide diversity and representative challenges. Our results highlight the limitations of state-of-the-art models when faced with real data and show simple baselines such as k-means or congealing can outperform more sophisticated approaches on such data.
Our dataset and code can be found at \href{https://printed-ornaments.github.io/}{https://printed-ornaments.github.io/}.

\end{abstract}
%
%
%

\section{Introduction}
\label{sec:intro}
\begin{figure}[t]
\centering
\subfloat[\centering Clustering]{\label{fig:img-clus}\includegraphics[width=0.95\textwidth]{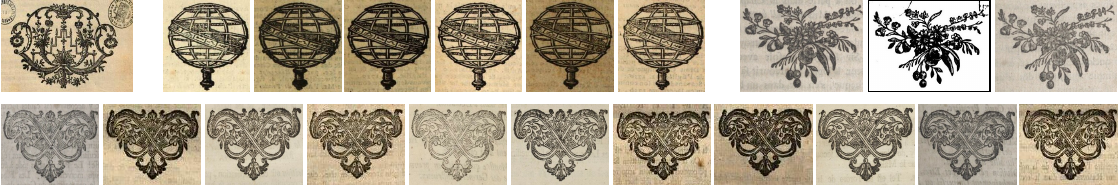}}\par
\begin{minipage}[b]{.45\linewidth} 
\centering
\subfloat[\centering Element discovery]{\label{fig:obj-dec}\includegraphics[width=\textwidth]{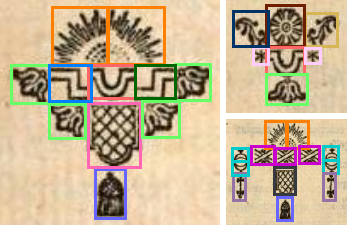}}
\end{minipage}
\begin{minipage}[b]{.45\linewidth}
\centering
\subfloat[\centering Unsupervised change localization]{\label{fig:vign}\includegraphics[width={\textwidth}]{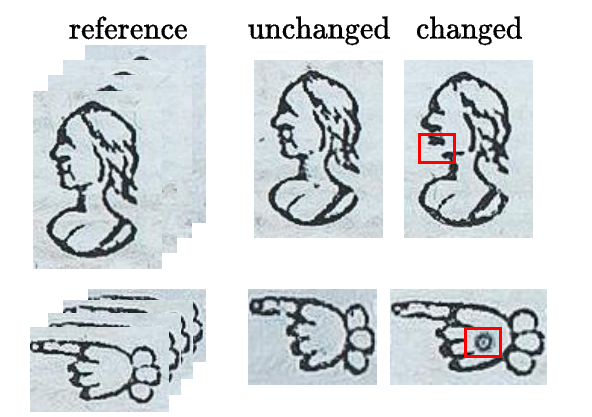}}
\end{minipage}
\caption{\textbf{Our Rey's Ornaments dataset.} Our dataset, based on ornaments found in the books published by or attributed to Marc-Michel Rey (1720-1780), focuses on three unsupervised computer vision tasks that are of interest to book historians: (a) image clustering of ornaments printed using wooden blocks, (b) unsupervised element discovery in composite ornaments printed using multiple types of vignettes, and (c) unsupervised change localization in vignette series.}
\vspace{-1em}
\label{fig:teaser}
\end{figure}

Typographical ornamentation is a key component of historical printed texts. While ornaments were first collected to improve the attribution of books to a particular printer~\cite{Corsini1988,Wilkinson2013,McKennaMori2019}, they turned out to be a critical part of material bibliography and book archaeology \cite{Riffaud2011}. Because of the massive amount of available material and the difficulty of annotations, many aspects of the study of book ornaments could benefit from modern unsupervised computer vision. In this paper, we identify three tasks of particular interest, illustrated in Fig.~\ref{fig:teaser}: clustering, element discovery, and unsupervised change localization. For these three tasks, we built datasets and evaluated state-of-the-art methods, showing that despite the apparent simplicity of the 2D patterns to analyze, these tasks remain extremely challenging and would benefit from more attention.

The challenges come from three main reasons: First, book ornaments are complex objects.
Older ornaments come from unique wooden blocks, that might have similar appearances but that historians want to differentiate. More recent ornaments are assembled from several vignettes produced by typographical metal types. Such composite ornaments may include tens of vignettes, which one would like to identify as different visual elements, and possibly relate to catalogs that were used to sell metal types. Second, the appearance of each ornament has a lot of variations. It can be due to many factors, including of course degradation of the books and image acquisition conditions, but also aging or degradation of the blocks, 3D effects, variations in the vignettes assembling, randomness in inking, in the transfer of the ink to different papers and the hand press inking process. Third, the tasks that are of actual interest to book historians are challenging in themselves: no supervision, very few samples corresponding to each ornament or vignette, highly imbalanced data, and the importance of fine differences.

To the best of our knowledge, there are no publicly available sets of printed ornaments annotated for ornament analysis tasks. We built our dataset based on the books published by Marc-Michel Rey (1720-1780), the leading publisher of Enlightenment philosophers, who is known to be especially attentive to the quality of his books \cite{Rey_database,BahierPorte2022}. The XVIIIth century is a particularly interesting period from the book-ornament point of view since it marks the transition between the dominant use of woodblocks and typographic metal types. Restricting ourselves to a single bookseller is motivated both by historical considerations, the history of this particular bookseller being of interest, and practical ones since this leads to a limited vocabulary of vignettes and woodblocks which makes annotation by historians possible. 

Our Rey's Ornaments dataset is composed of three parts, giving insights into our three tasks, each one composed of distinct image sets: First, our clustering dataset, based on woodblock ornaments, includes 167 images of 36 different ornaments, each associated with 3 to 14 occurrences. We found that DTI Clustering~\cite{monnier_deep_2020} outperformed state-of-the-art clustering approaches by a large margin, but that the k-means algorithm~\cite{macqueen1967some} on foundation features (e.g., CLIP~\cite{radford2021learning}) performed almost on-par when classes are very imbalanced. However, all algorithms led to less than 80\% accuracy in this setting.  Second, our element discovery dataset includes 100 images of composite ornaments containing a total of 1271 elements, from a dictionary of 72 different vignettes. We found all unsupervised element discovery methods to perform poorly. We believe this stresses the potential benefit of our dataset for the community, compared to the synthetic datasets often used to evaluate unsupervised object segmentation methods.  Third, our unsupervised change localization dataset includes 30 types of vignettes with four reference instances and two test instances, a normal one and one where changes have been annotated. We found that reconstruction-based approaches, such as congealing or VAE-based methods, performed poorly compared to human annotations, mostly because they are confused by the variations in inking. We see this as an invitation to better formalize the notion of changes relevant to book historians and design-associated algorithms.

The paper is organized as follows: Section \ref{sec:relatedwork} reviews related work for databases and our three tasks, Section \ref{sec:dataset} presents our Rey's Ornaments dataset, tasks, and metrics, and Section \ref{sec:results} discusses the performance of state-of-the-art algorithms for each task. 

\section{Related work}
\label{sec:relatedwork}

\paragraph{Ornaments repositories.}

Several large repositories of ornaments used to decorate printed books in the hand-press period (1440-1830) exist, initially motivated by the identification of printers \cite{corsini1999}.  The XVIIIth century marks the transition from woodblocks to typographic metal types.
Such repositories include the Fleuron database~\cite{Fleuron} (several thousands of XVIth-XVIIIth century ornaments printed in French-speaking Switzerland), the Maguelone database~\cite{Maguelone} (over 7,000 XVIIIth century ornaments printed in France and Europe), the Broadside ballads database~\cite{BodleianBallads} (centered on woodblocks extracted from English broadside ballads), and the Compositor database~\cite{Compositor} (including over a million ornaments extracted from XVIIIth century English books~\cite{Wilkinson2021}). The need for image retrieval tools for these catalogs emerged early~\cite{Bigun1996,Corsini2001}, many have been developed specifically for ornaments~\cite{Baudrier2009}, and some are available directly with the database.
A popular, and more general, alternative is the Visual Image Search Engine (VISE)~\cite{Dutta2021,bergel2013content,Chung2014}. Automatic tools, however, rarely go beyond image retrieval, and these databases do not provide the detailed annotations necessary to evaluate more advanced algorithms. 

\paragraph{Image clustering.} 
The classical k-means algorithm \cite{macqueen1967some} remains the basis for many recent clustering methods. It splits a collection of images into k clusters by jointly optimizing k centroids and the assignment of each image to the closest centroid. Distances between images can be directly computed in pixel space, for example in Transformation Invariant methods~ \cite{frey2001flti,frey2003tiem,monnier_deep_2020}. More commonly, k-means is used with learned features, either optimized for a pretext task or together with the k-means clustering.
For example, DCN~\cite{yang2017towards} optimizes an auto-encoder both for reconstruction and clustering in latent space, CCNN~\cite{hsu2017cnn} fine-tunes a pre-trained network to minimize a mini-batch k-means loss, and DeepCluster~\cite{caron_deep_2018} learns features in a self-supervised way using k-means clusters as class labels.
There are of course many different clustering methods. Recently, many approaches defined a loss that is simply minimized with stochastic gradient descent in a deep learning framework. In some, clustering and feature learning are optimized jointly, for example with a loss based on KL divergence~\cite{xie2016unsupervised,ijcai2017p243}, mutual information~\cite{hu2017learning,ji2019invariant}, consensus~\cite{Huang2022DeepCluEEI,metaxas2023divclust}, or image likelihood~\cite{kosiorek_stacked_2019}. This can also be done with two-step approaches, first learning features and then optimizing a clustering objective on these features, for example in SCAN~\cite{vangansbeke2020scan}.

Although clustering methods are adopted in various fields in cultural heritage \cite{mlch2020}, especially in document analysis, the focus mainly remained on texts rather than visual clues \cite{He2016AMG,Goyal2020APG}. We evaluate different types of approaches for our dataset: k-means on pixels, transformation invariant k-means~\cite{monnier_deep_2020}, k-means on pre-trained features, joint feature and cluster learning with mutual information~\cite{ji2019invariant} and consensus objectives~\cite{metaxas2023divclust}, as well as optimizing a clustering objective on self-supervised features~\cite{vangansbeke2020scan}.

\paragraph{Element discovery.}  By element discovery, we mean identifying different categories of visual elements and decomposing images into such elements without supervision. This problem is related to object co-segmentation and discovery, which has been addressed for example by using visual words and a topic hierarchy~\cite{sivic2008unsupervised,russellUsingMultipleSegmentations2006,caoSpatiallyCoherentLatent2007} or by computing similarities between image regions~\cite{graumanUnsupervisedLearningCategories2006,joulinDiscriminativeClusteringImage2010,rubinsteinUnsupervisedJointObject2013,choUnsupervisedObjectDiscovery2015,voUnsupervisedImageMatching2019,shen_discovering_2019}.  However such methods are designed to work with textured and discriminative regions and are unlikely to work with our composite ornaments.

Element discovery is also related to what has recently been referred to in deep learning as unsupervised multi-object segmentation. However, these methods do not always model background and often do not model any notion of class or category for the discovered segments. We give an overview of these models following the classification introduced in~\cite{karazija2021clevrtex}. 
{\it Pixel-space approaches}, such as~\cite{burgess_monet_2019,engelcke2021genesis,engelcke_genesis_2020}, model images using a predefined number of objects, and determine per-pixel allocations to objects without discriminating two occurrences of the same object.
An additional limitation is that these approaches are typically computationally intensive and restricted to a limited number of objects.
Thus we do not evaluate them, but some of their core principles are integrated into the methods presented below. 
{\it Glimpse-based approaches}~\cite{eslami2016attend,crawford2019spatially,Lin2020SPACE,jiang2020generative,zhu2021gmair,sauvalle2023unsupervised} follow the seminal AIR \cite{eslami2016attend} method and perform element prediction based on regions from the input image, referred to as `glimpses'. 
We evaluated the SPACE~\cite{Lin2020SPACE} and AST-argmax~\cite{sauvalle2023unsupervised} approaches, which have the advantage of having a background model. However they do not differentiate classes of elements, thus we combine them with feature clustering to obtain categories on the extracted elements. To the best of our knowledge, the only method in this category that identifies classes is GMAIR~\cite{zhu2021gmair}. We thus also evaluate it, together with the SPAIR~\cite{crawford2019spatially} method it builds on, but both methods suffer from not having a background model.
{\it Sprite-based approaches}~\cite{smirnov_marionette_2021,monnier_unsupervised_2021} decompose images into elements by learning representative prototypes and their transformation to optimize a reconstruction loss. We evaluate the DTI-Sprites approach~\cite{monnier_unsupervised_2021}, because it models the scale of visual elements, which is critical for our data.

These unsupervised multi-object segmentation methods are mostly evaluated quantitatively only on synthetic datasets such as Tetrominoes~\cite{greffMultiObjectRepresentationLearning2019}, CLEVR6~\cite{johnsonCLEVRDiagnosticDataset2017,greffMultiObjectRepresentationLearning2019}, multi-dSprites~\cite{multiobjectdatasets19}, or ClevrTex~\cite{karazija2021clevrtex}, where performances are typically very good. However, they have been shown to perform very poorly on natural 3D world images~\cite{yang2022promising}. Because real applications are unclear, the evaluations often focus mainly on instance segmentation, i.e., not considering semantics. Thus, we believe the introduction of a real dataset, challenging but simpler than natural scenes, with a clear associated task, can be a significant contribution to this entire field and improve the benchmarking of future methods.

\paragraph{Unsupervised change localization.}  

By unsupervised change localization, we refer to methods that identify pixels where changes occur in a test image compared to a reference, which can also be referred to as anomaly or novelty detection in the literature. The reference can be a single image sample or a small collection of samples. Since many methods address this task and a complete review is out of the scope of this work, we refer the reader to existing reviews and benchmarks~\cite{pimentel2014review,bergmann2019mvtec,pang2021deep,ruff2021unifying}. Approaches can be broadly separated between reconstruction- and feature-based methods. To localize anomalies or changes, the reconstruction-based methods compute a pixel-wise difference between the reconstructed image and input. They usually need class-dependent thresholds and anomalous training images. This motivated the development of feature-based methods such as \cite{venkataramanan2020attention}. Such methods compute a false-color heatmap from the extracted features to highlight regions containing (a apart of) the anomalies \cite{liu2020towards}. Feature-based methods typically do not precisely localize the changed pixels; thus we focus on reconstruction-based methods.

The reconstruction can be either directly computed from the reference samples or produced by a deep neural network. We consider as baseline simply comparing the test image to the average of the reference images, or solving for the joint alignment problem before computing the average, which can be seen as a variation of the classical congealing problem~\cite{learned2005data,cox2008LScongealing}. 
Deep reconstruction approaches are typically based either on GANs~\cite{schlegl2017unsupervised,zenati2018efficient,akcay2019ganomaly} or auto-encoders~\cite{kingma2013auto,baur2019deep}, which in theory can capture complex variability in the reference examples. We benchmark both a classical VAE-based method~\cite{kingma2013auto} and a more recent auto-encoder-based method that leverages spatial transformers~\cite{jaderberg_spatial_2015} to obtain sharper reconstructions~\cite{chaki2023STAE}.

Change localization datasets exist in various domains such as 3D-MR-MS \cite{lesjak2018novel} in medical imaging, S2Looking \cite{shen2021s2looking} in remote sensing, CDnet \cite{goyette2012changedetection}  in video surveillance, MVTec \cite{bergmann2019mvtec} in industrial inspection, and TAMPAR \cite{naumann2024tampar} in tampering detection. To the best of our knowledge, no such dataset exists for historical ornaments, while change localization is one of the key steps in analyzing printed material, and inking variability makes this task very specific. 

\section{The Rey's Ornaments dataset}
\label{sec:dataset}

\begin{table}[t!]
    \renewcommand{\arraystretch}{1}
    \centering
    \begin{subtable}[t!]{\textwidth}
    \centering
    \resizebox{0.43\linewidth}{!}{
    \begin{tabular}{@{}l r r r@{}}
    \toprule
    Clustering & Base & Imbalanced & Balanced \\
    \midrule
    \# ornaments & 339 & 167 & 70 \\
    \# classes & 163 & 36 & 14 \\
    \bottomrule
    \end{tabular}}
    \hspace{0.1cm}
    \raisebox{-.4\height}{
    \includegraphics[width=0.4\textwidth,trim={20 10 0 100},clip]{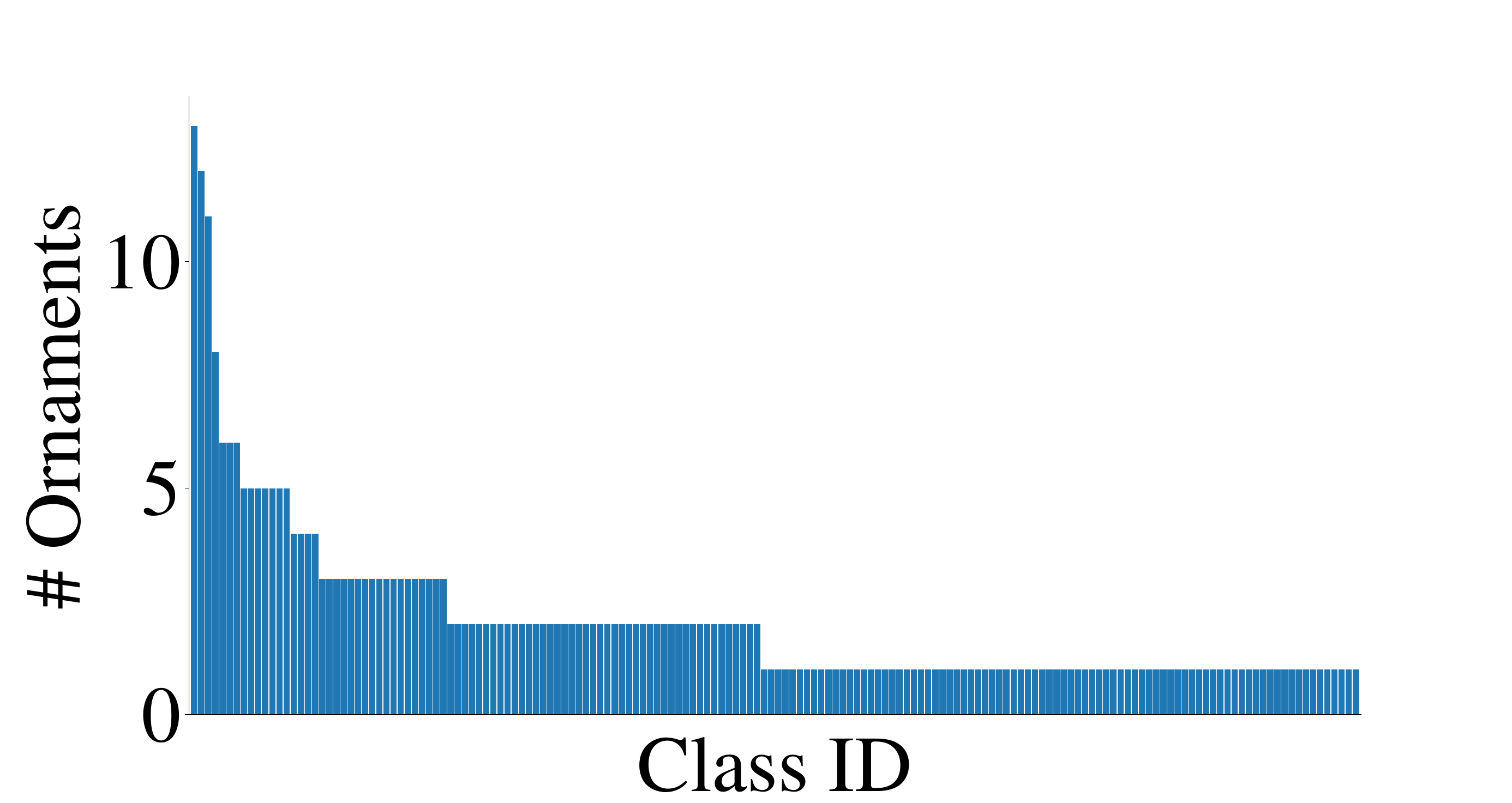}}
    \caption{Clustering dataset}
    \label{fig:clustering_stat}
    \end{subtable}
    \begin{subtable}[t!]{\textwidth}
        \centering
    \resizebox{0.2\linewidth}{!}{
    \begin{tabular}{@{}l r@{}}
    \toprule
    \multicolumn{2}{c}{Element Discovery} \\
    \midrule
    \# {comp. ornaments} & 100 \\
    \# {vignettes} & 72 \\
    \# vignettes occ. & 1271 \\
    \bottomrule
    \end{tabular}}~~
     \hspace{0.1cm}   
    \raisebox{-.45\height}{
    \includegraphics[width=0.35\textwidth,trim={0 0 100 50},clip]{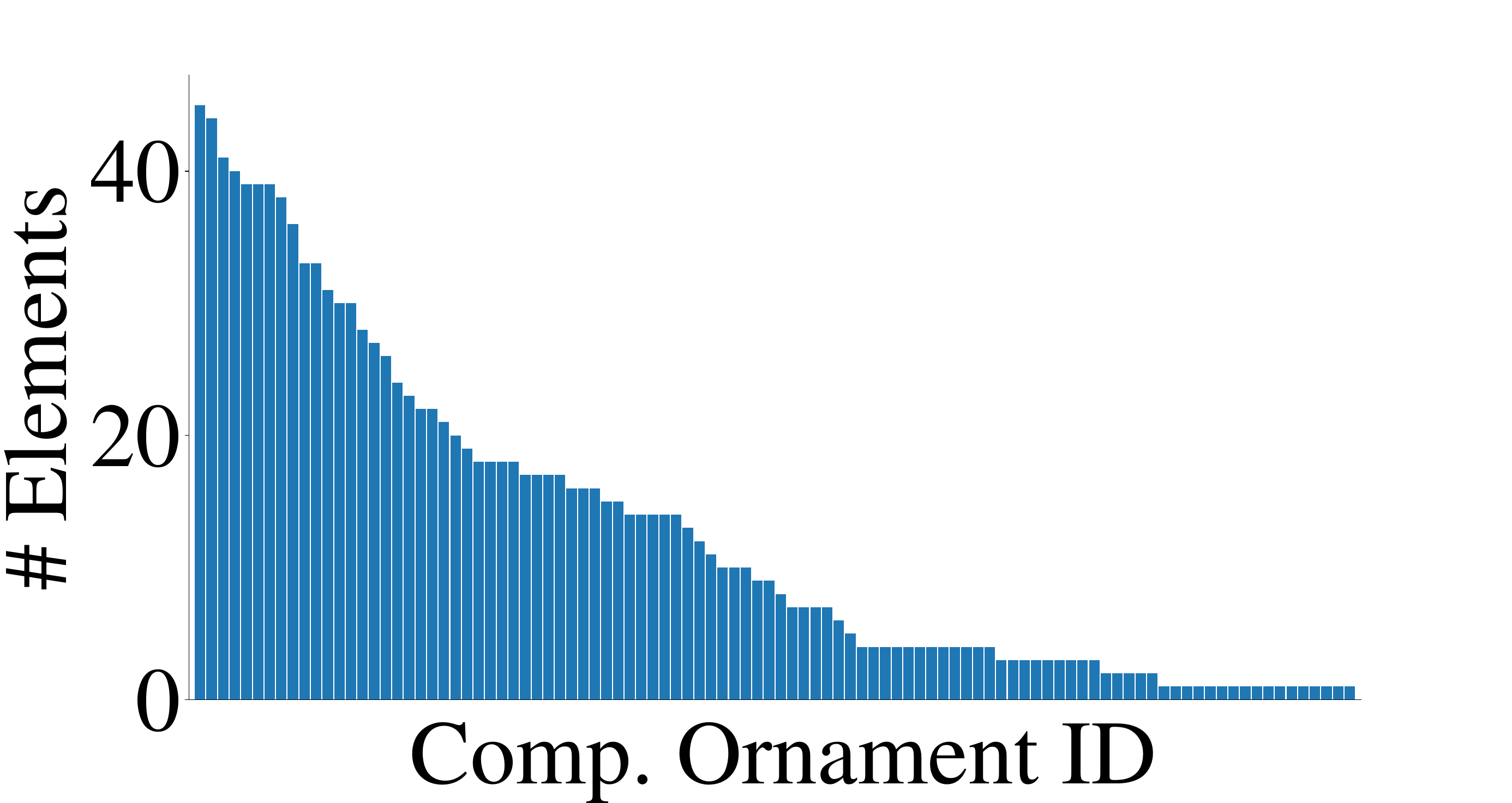}}~~
    \raisebox{-.43\height}{
    \includegraphics[width=0.35\textwidth,trim={0 0 100 50},clip]{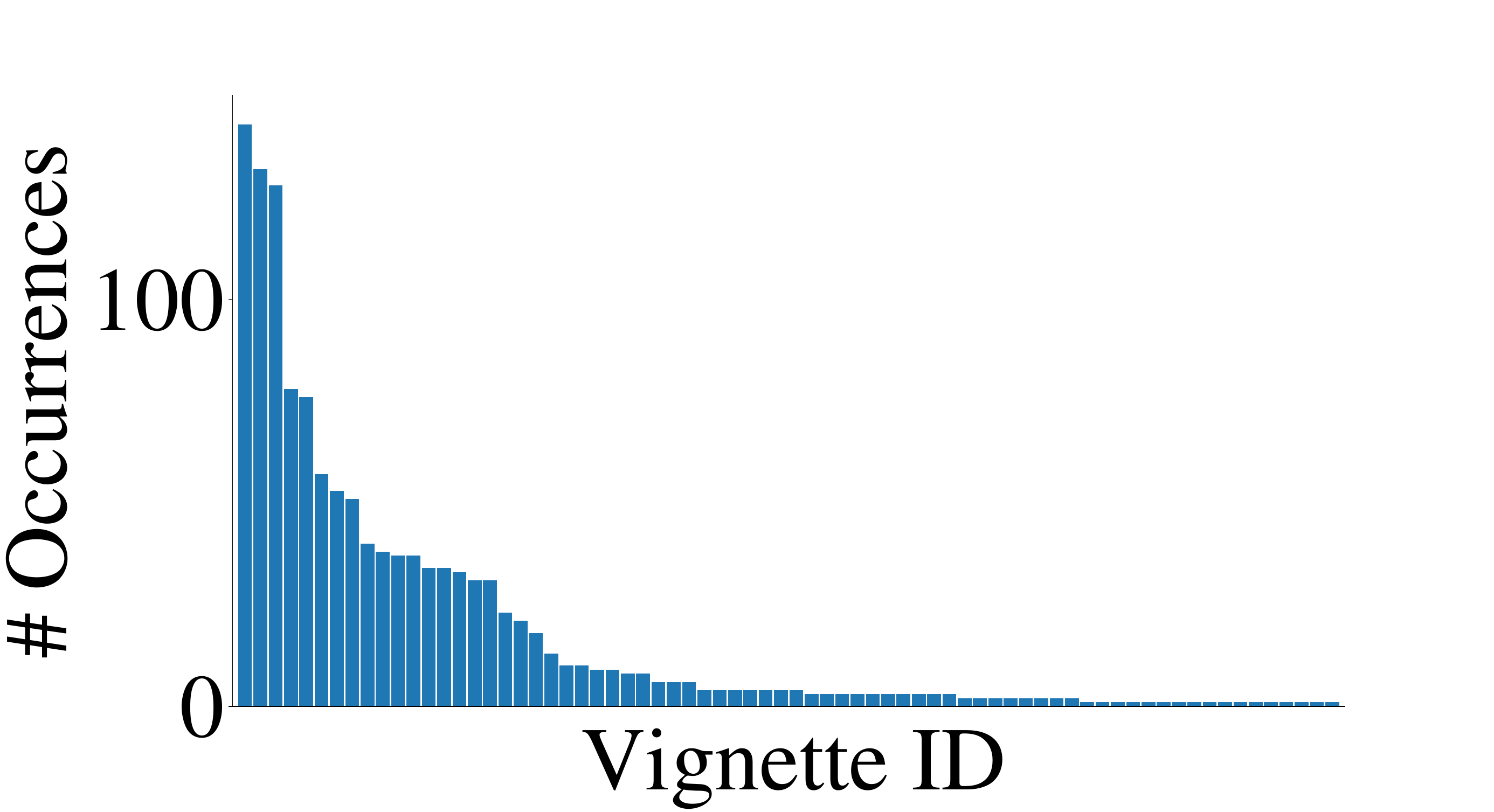}}
    \centering
    \caption{Element discovery dataset}
    \label{fig:discovery_stat}
    \end{subtable}
    \label{tab:data-stats}
    \caption{\textbf{Dataset statistics.} Our clustering and element discovery datasets are highly imbalanced, which is one of the  key challenges of real data but rarely considered in benchmarks.}
    \vspace{-1em}
\end{table}
Our dataset is composed of blocks and composites but focused on a bookseller based in Amsterdam: Marc-Michel Rey (1720-1780). Here the motivation is the study of editorial practices during a given censorship period. It leads to the problem of book attribution for which the study of Rey's correspondences \cite{BahierPorte2021} is helpful. Automatically identifying the wooden blocks used by Marc-Michel Rey and finding a composition style in the composites represent new challenges, in a well-circumscribed corpus with well-documented metadata including positions in the books and geographical location of the volumes. This results in a limited vocabulary of wooden blocks and vignettes which makes annotations possible on several hundred ornaments.

This section presents the three parts of the dataset, targeted toward clustering, element discovery, and unsupervised change localization. The ornaments for each task are extracted from books listed in the Marc-Michel Rey database~\cite{Rey_database} and shared by the Bibliothèque Nationale de France (BnF) or the Bibliothèque Municipale de Lyon (BML). All the annotations were made under the supervision of book historians.
\subsection{Block ornaments and clustering}

The clustering dataset contains ornaments printed from wooden blocks. Examples are shown in Figure~\ref{fig:img-clus}. We considered an initial set of 339 images of block ornaments, which we refer to as the base set, and annotated their class labels with the help of VISE~\cite{Dutta2021}. This led to 163 ornament classes, most of them corresponding to one or two images. While we will release this full set and the associated annotations, we focused our evaluation on two subsets:
\begin{itemize}
    \item an imbalanced subset, with all the 167 images from the 36 classes that have at least 3 instances, 
    \item a balanced subset, with 70 images, built by randomly sampling 5 images from the 14 classes that have at least 5 instances.
\end{itemize}
This choice was motivated by the fact that most existing clustering approaches do not handle well a large number of classes with a single or very few samples. The statistics of our dataset are visualized in Figure~\ref{fig:clustering_stat}.

\subsubsection{Evaluation metrics. }
We use two standard metrics to evaluate our clustering: normalized mutual information (NMI) and accuracy. Following the standard practice to evaluate clustering, we compute accuracy by matching clusters and classes using the Hungarian matching algorithm~\cite{Kuhn1955TheHM}.

\subsection{Composite ornaments and element discovery} 
The element discovery dataset contains 100 images of composite ornaments. In this set, we first identified 72 categories of vignettes as composition elements. We identified 51 of those in two vignette catalogs by providers of metal types used in Rey's publishing~\cite{Enschede,Rosart}.
For each category of vignettes we selected a representative example that we did not use in our experiments but was the reference for annotation and that we will release as part of the dataset.
We then manually annotated the bounding box and class of each element in each composite ornament. Three examples of composite ornaments annotated with semantic bounding boxes for each {vignette} can be seen in Figure~\ref{fig:obj-dec}.  The statistics of the resulting dataset are presented in Table~\ref{fig:discovery_stat}. Our 100 composite ornaments contain a total of 1271 elements, each corresponding to one of the 72 vignettes. Some ornaments included a single element, others more than 40. Some vignettes are used a single time, others more than 100 times. This strong imbalance is a part of the challenge of our dataset and representative of statistics encountered in real problems. Another challenge is that the elements are often grouped, very close, or even touching each other, and are thus much more challenging to separate than the objects in the synthetic unsupervised object segmentation datasets~\cite{greffMultiObjectRepresentationLearning2019,johnsonCLEVRDiagnosticDataset2017,multiobjectdatasets19,karazija2021clevrtex}.
\begin{figure}[t]
    \centering
    \includegraphics[scale=0.4]{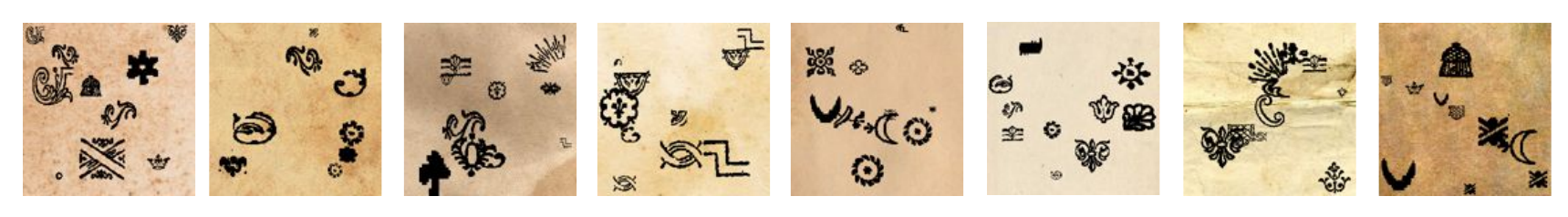}
    \caption{Examples of composite {ornaments} from our synthetic dataset.}
    \label{fig:ScatteredVignettesExample}
    \vspace{-1em}
\end{figure}

\subsubsection{Synthetic dataset.}
Because we found that the existing algorithms trained directly on our dataset performed very poorly, we also created a synthetic dataset, more similar to the ones existing in the literature. We first selected 67 distinct parchment images as empty backgrounds, then pasted on these backgrounds up to 10 elements uniformly sampled from our dictionary at random locations. Examples of the resulting synthetic composites are shown in Figure~\ref{fig:ScatteredVignettesExample}. We used these simpler examples, without annotations, to pre-train algorithms, and then fine-tuned them on the real data, which resulted in improved performances (see Section~\ref{sec:expe-disc}).

\subsubsection{Evaluation metrics. }
Recent unsupervised multi-object segmentation methods typically focus their evaluations on {\it instance} segmentation, often discarding background pixels. However, we want to identify the classes of the elements as well as their localization. Coarsely localizing them is also sufficient for applications, while annotating exact segmentation would be costly. We thus turn to an object detection metric, namely mean average precision (mAP). Following standard practices~\cite{pascal}, we consider a detection to be accurate if its Intersection over Union (IoU) with the annotation is larger than 0.5. We follow the same approach as for clustering algorithms evaluation and use the Hungarian matching algorithm~\cite{Kuhn1955TheHM} to match the discovered categories of elements with vignette categories, similar to~\cite{simeoni2021lost}. To analyze separately the influence of the elements' localization and identification, we also measure class-independent element detection and report the results using average precision (AP). 
\subsection{Vignettes and unsupervised change localization}

The change localization dataset contains 180 images of elements printed in vignettes' catalogs published by Enschede~\cite{Enschede} and Rosart~\cite{Rosart}, two providers of metal types used by Rey to build composite ornaments, as well as in the catalog published by Fournier~\cite{Fournier} which was used by Rey's counterfeiters. In such catalogs, the same vignette is printed several times on one or multiple lines. This enabled the publishers to see variations that could be expected when hand-press printing a given vignette. These catalogs enabled us to easily find prints perceived as different from the other ones for the same vignette.
We built a dataset of 30 vignette types, each associated with 6 images: 4 images of standard prints we used as reference, and 2 test images, one associated with a 5th standard print labeled as `unchanged' and one associated with an error print labeled as `changed'. Examples of this dataset are shown in Figure~\ref{fig:vign}. We annotated the binary masks and the bounding boxes  of the pixel-wise regions corresponding to the perceived changes.

\subsubsection{Evaluation metrics. } We evaluate change localization by computing Intersection over Union (IoU) between annotated and predicted changes on each test image. We compute this metric either only on the changed image or on both the changed and unchanged images - which is harder since the method has to correctly recognize that none of the variations in the unchanged images are significant. We then report a mean IoU (mIoU) over the 30 vignette classes.

\begin{table}[t!]
    \centering
    \renewcommand{\arraystretch}{1}
    \begin{tabular}{@{}l l c c c c@{}}
        \toprule
        \multicolumn{2}{l}{Dataset} & \multicolumn{2}{c}{Imbalanced Dataset} & \multicolumn{2}{c}{Balanced Dataset} \\
         \midrule
         \multicolumn{2}{l}{Method} & Acc. (\%) $\uparrow$ & NMI (\%) $\uparrow$ & Acc. (\%) $\uparrow$ & NMI (\%) $\uparrow$ \\
         \midrule
         \multicolumn{2}{l}{\textit{clustering over features}} & & & & \\
         \multicolumn{2}{l}{IIC~\cite{ji2019invariant}} & 19.2 & 40.6 & 25.0 & 35.0 \\
         \multicolumn{2}{l}{SCAN~\cite{vangansbeke2020scan}} & 46.1 & 68.7 & 47.1 & 64.6 \\
         \multicolumn{2}{l}{DivClust~\cite{metaxas2023divclust}} & 54.1±0.5 & 80.5±0.3 & 67.0±2.1 & 78.6±0.7 \\
         & \underline{Feat. Extractor} & & & & \\
         \multirow{2}{*}{k-means~\cite{macqueen1967some}} & DINO-ViTB16\cite{caron2021emerging} & 73.9±0.9 & \underline{89.9±0.4} & 70.9±3.3 & 86.8±1.6 \\
         & CLIP-RN50x16\cite{radford2021learning} & \underline{74.6±3.0} & \underline{90.1±1.1} & 78.3±2.9 & 89.8±1.2 \\
         \midrule
         \multicolumn{2}{l}{\textit{clustering over pixels}} & & & & \\
         \multicolumn{2}{l}{k-means~\cite{macqueen1967some}} & 65.6±1.8 & 84.1±1.1 & 74.3±1.9 & 83.9±0.7 \\
         \multicolumn{2}{l}{DTI Clustering~\cite{monnier_deep_2020}}
         & \textbf{75.7±0.8} & \textbf{90.7±0.6} & 
\textbf{87.4±1.9} & \textbf{92.1±1.2} \\
         \bottomrule
    \end{tabular}
    \vspace{0.8em}
    \caption{\textbf{Image clustering baselines.} We report the clustering accuracy (Acc.) and normalized mutual information score (NMI) on our imbalanced and balanced datasets. We report the standard error among 5 runs for the fastest methods, outline the best results in bold, and underline ones within one standard error.}
    \vspace{-1em}
    \label{tab:clus_baseline}
\end{table}

\section{Results and analysis}
\label{sec:results}
In this section, we  analyze the results of diverse methods for each of our tasks. This provides insight both into the specific challenges of our problems, and the strengths and limitations of state-of-the-art algorithms.

\subsection{Block ornaments and clustering}
 \label{sec:baseline_clus}
\subsubsection{Methods} We tested both methods that performed clustering on pixels and features. Using the raw pixel values and the standard L2 distance, we evaluated k-means \cite{macqueen1967some}, and DTI Clustering \cite{monnier_deep_2020} which jointly learns and transforms cluster centers to reconstruct images. 
For clustering on features, we used both pre-trained standard features and methods that specifically learn features for clustering on a specific dataset. As examples of standard pre-trained features, we used the self-supervised DINO~\cite{caron2021emerging} features and the supervised CLIP~\cite{radford2021learning} features. As methods that jointly train for features and clustering, we evaluated IIC~\cite{ji2019invariant}, SCAN~\cite{vangansbeke2020scan}, and DivClust~\cite{metaxas2023divclust}.

\begin{figure}[t!]
    \centering
    \resizebox{\linewidth}{!}{
    \begin{tabular}{c}
         \includegraphics[width=\textwidth]{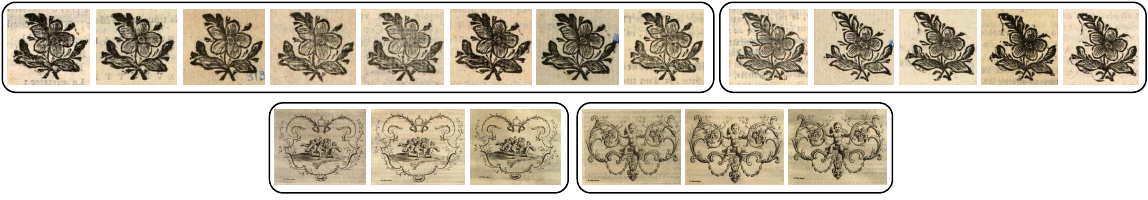}\\
         (a) {Valid clusters obtained with DTI Clustering.} \\
         \includegraphics[width=\textwidth]{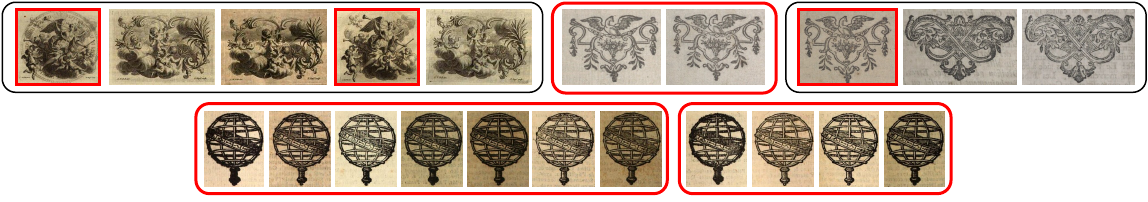}\\
         (b) {Failure examples obtained with DTI Clustering.}
    \end{tabular}
    }
    \caption{\textbf{Qualitative results for clustering.} Although clusters obtained with DTI Clustering are often valid (a), there are failure cases (b) due to e.g. similar vignettes (top), or split clusters (bottom). Results are qualitatively similar when using k-means with pre-trained feature extractors.} 
    \vspace{-1em}
    \label{fig:img-clus-qual}
\end{figure} 

\subsubsection{Results}
Our quantitative results are reported in Table \ref{tab:clus_baseline} and highlight several facts. First and most surprisingly, simply performing k-means on pixel values performed better than methods learning ad-hoc features and clustering on both the balanced and imbalanced datasets. We believe that can be explained by the fact that our images are relatively aligned and similar in appearance (dark ink on light paper), and by the fact that our datasets have very few examples per class. This highlights a clear limitation of common benchmarks and the complex methods that report state-of-the-art performances on them. Second, performing k-means over pre-trained features performs better than any other feature-based method, in particular for the imbalanced dataset, with a small advantage for the CLIP features compared to the DINO features. Third, DTI Clustering consistently improves over all approaches, but the margin is small on the imbalanced dataset.

Figure~\ref{fig:img-clus-qual} shows some qualitative results obtained with the best-performing method, DTI Clustering, on the imbalanced dataset. Valid clusters are obtained even with variations in inking and paper appearance (Fig.~\ref{fig:img-clus-qual}a). Failure cases are typically related to similar-looking but different ornaments being grouped in the same cluster, or different versions of the same ornament being split into two clusters (Fig.~\ref{fig:img-clus-qual}b). 

Altogether, we find it both surprising and interesting to see that no standard method ables to perfectly solve the simple task of aligned, printed patterns' clustering and believe that it demonstrates the interest of our dataset to evaluate and help design new algorithms.

\subsection{Composite ornaments and element discovery} 
\label{sec:expe-disc}

\subsubsection{Methods}
As explained in Section~\ref{sec:relatedwork}, we focused on unsupervised instance segmentation approaches and evaluated the SPACE~\cite{Lin2020SPACE}, AST-argmax~\cite{sauvalle2023unsupervised}, SPAIR ~\cite{crawford2019spatially}, GMAIR~\cite{zhu2021gmair} and DTI-Sprites~\cite{monnier_unsupervised_2021} approaches. Only GMAIR and DTI-Sprites provide categories for the different elements. Thus, we used clustering on the discovered element regions to obtain categories and compute mean average precision, that corresponds to the task that would make the most sense for applications. We do this clustering using k-means on CLIP features.

\begin{table}[t!]
\centering
\renewcommand{\arraystretch}{1}
\begin{tabular}{@{}lC{3em}C{6em}C{6em}C{6em}C{6em}@{}}
\toprule
&\textbf{} &\multicolumn{2}{c}{\textit{Real training data} }& \multicolumn{2}{c}{\textit{Synt. training data} }\\ 
\textbf{Model} & \textbf{Bkg} &AP(\%) $\uparrow$& mAP(\%) $\uparrow$& AP(\%) $\uparrow$ & mAP(\%) $\uparrow$ \\ 
\midrule
SPAIR~\cite{crawford2019spatially} &\xmark&0&0&0&0\\
GMAIR~\cite{zhu2021gmair} &\xmark&0&0&0&0\\
SPACE~\cite{Lin2020SPACE} &\cmark &0&0&8.1&6.1\\
AST-argmax~\cite{sauvalle2023unsupervised} &\cmark&13.6&13.2&\textbf{38.4}&\textbf{27.6} \\
DTI-Sprites~\cite{monnier_unsupervised_2021} &\cmark&0&0&0&0 \\ 
\bottomrule
\end{tabular}
\vspace{0.8em}
\caption{{\bf Quantitative results for element discovery}. For each method, we report results both for models trained directly on real composite ornaments, and pre-trained on synthetic and fine-tuned on real composite ornaments. We report category agnostic average precision (AP) and mean average precision (mAP) using the categories obtained with k-means and Hungarian matching over CLIP features of discovered elements.}
\vspace{-1em}
\label{tab:element_discovery_comparison}
\end{table}

\begin{figure}[htbp!]
\centering
\resizebox{\linewidth}{!}{
    \begin{tabular}{ccccccc}
        \inputlinedecomp{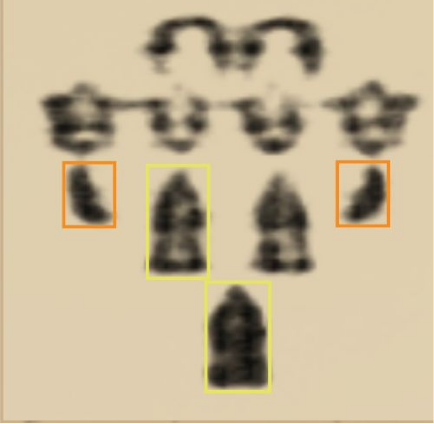}
        \inputlinedecomp{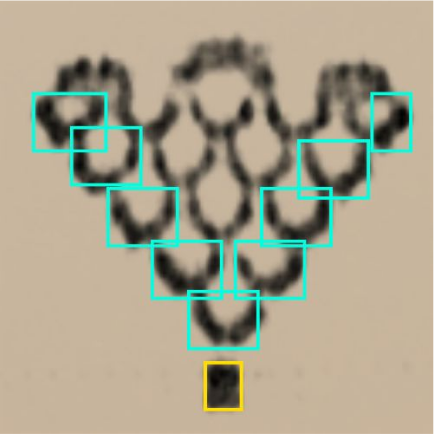}
        \inputlinedecomp{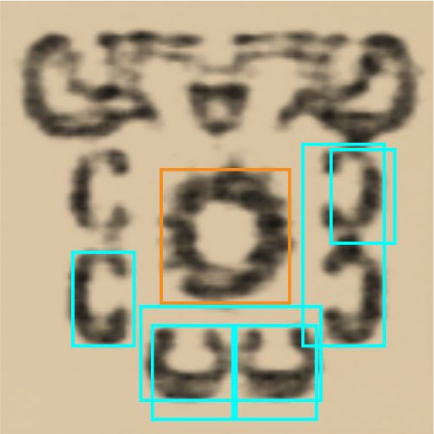}
        \inputlinedecomp{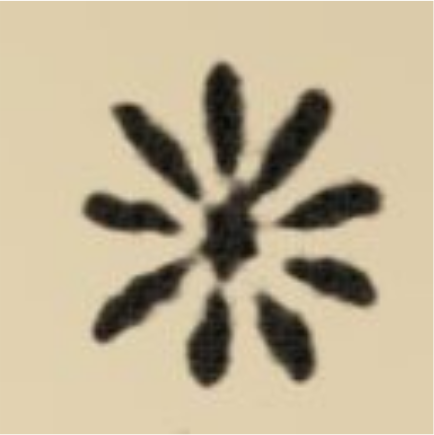}
        \inputlinedecomp{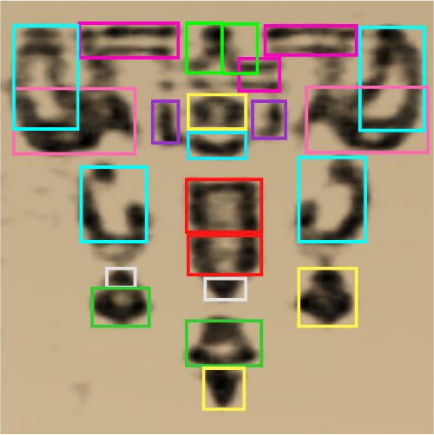}
        \inputlinedecomp{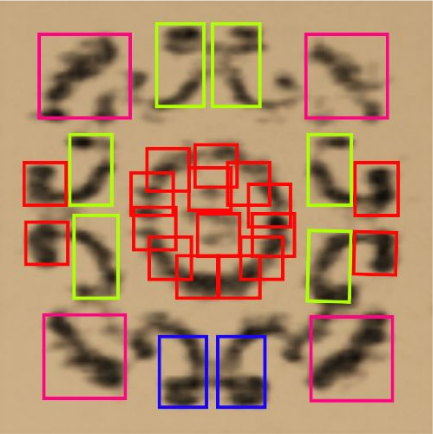}
        \inputlinedecomp{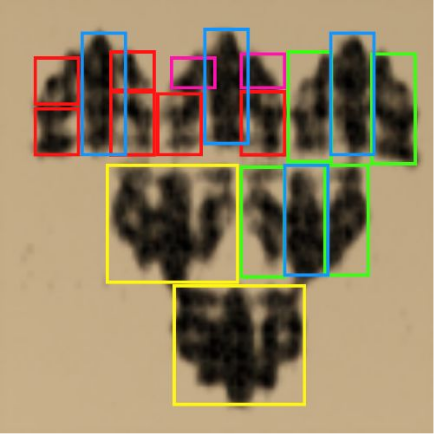}
        \inputlinedecomp{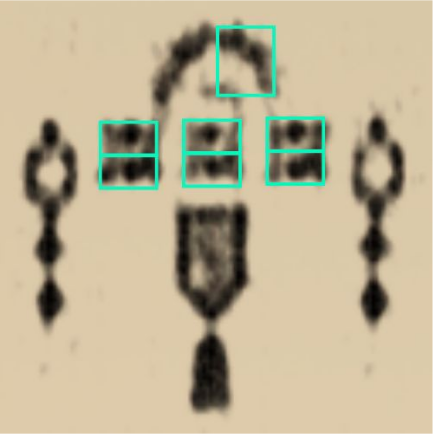}
         (a) Original & (b) GT & (c) SPAIR & (d) GMAIR & (e) SPACE & (f) AST & (g) DTI\\ 
    \end{tabular}
    }
    \caption{{\bf Qualitative results for element discovery.} We show the dataset images (a) with their semantic ground truth bounding boxes (b) and the reconstruction and predicted semantic bounding boxed from different models (c-g). For all the methods, we show the results of the models pre-trained on the synthetic dataset and fine-tuned on the real dataset, and the semantic boxes obtained using k-means on CLIP features.}
    \label{fig:decomposition_qualitative}
\end{figure}

\subsubsection{Results}
Quantitative results are reported in Table~\ref{tab:element_discovery_comparison}. The most striking fact is that all methods perform extremely poorly when trained on our real dataset. A single method, AST-argmax~\cite{sauvalle2023unsupervised} produces meaningful detection results, and it only has $13.6\%$ average precision (AP).
We believe this is mainly due to two factors: first, the intrinsic complexity of the composite ornaments, where the different elements are often intentionally placed together and even connected to form more complex geometric patterns; and second, the difficult statistics of the dataset, with many rare vignettes, and ornaments composed of many elements. This was the motivation to pre-train our methods on our synthetic dataset, where the elements are sampled from the vignette dictionary uniformly and positioned randomly, making the elements easier to learn for the models. While this helps methods identify the elements on the synthetic dataset, and then boost performances when fine-tuning the models on real data, the performance remains relatively low. The qualitative results, shown in Figure \ref{fig:decomposition_qualitative}, give more insights into the reasons for these poor performances. Methods that do not incorporate a background model (SPAIR and GMAIR) provide good reconstructions but use many elements that seem randomly localized. On the contrary, glimpse-base methods that incorporate a background model (SPACE and AST-argmax) reconstruct part of the ornaments using their background model. However, they still manage to accurately detect some of the elements, with a clear improvement over training only with real data, even if the performance remains low ($38.4\%$ AP for the best method). Finally, we found that DTI-Sprites struggled to accurately reconstruct the data and thus produced irrelevant decomposition.

Note that the two methods that provided the best results, SPACE and AST-argmax, do not learn a class model and thus require clustering the detected patches and assigning clusters to classes using the Hungarian matching algorithm to compute the mean average precision (mAP). 

Note that the limitations we point out are in line with the conclusion of a recent study on the potential of unsupervised object segmentation in real-world images~\cite{yang2022promising}.
\begin{table}[t]
    \centering
    \begin{subtable}[b]{0.47\textwidth}
        \renewcommand{\arraystretch}{1}
        \resizebox{\linewidth}{!}{
        \begin{tabular}{lcccccccc}
            \toprule
            Method & \multicolumn{2}{c}{Naive} & \multicolumn{2}{c}{VAE~\cite{kingma2013auto}} & \multicolumn{2}{c}{STAE~\cite{chaki2023STAE}} & \multicolumn{2}{c}{Cong.~\cite{cox2008LScongealing}}\\
            \midrule
            Category & \gc C & CU & \gc C & CU & \gc C & CU & \gc C & CU\\
            \midrule
            dot1    & \gc0.0  & 0.0  & \gc2.5 & 1.8  & \gc5.2 & 2.5 & \gc\textbf{7.5} & \textbf{5.0} \\
            dot2    & \gc18.0 & 8.9  & \gc9.7 & 7.3 & \gc14.7 & 0.7 & \gc\textbf{38.5} & \textbf{35.7}\\
            dot3    & \gc19.8 & 7.5  & \gc18.2 & 9.3 & \gc23.4 & 5.7 & \gc\textbf{36.4} & \textbf{28.4}\\
            dot4    & \gc1.6  & 0.0  & \gc\textbf{5.0} & \textbf{5.4} & \gc4.7 & 1.2 & \gc0.0 & 0.0\\
            dot5    & \gc18.2 & 20.0 & \gc16.7 & 14.5 & \gc\textbf{24.5} & \textbf{31.3} & \gc21.8 & 23.5\\
            emblem1 & \gc27.1 & 21.7 & \gc22.8 & 16.3 & \gc2.2 & 2.5 & \gc\textbf{32.1} & \textbf{26.3}\\
            emblem2 & \gc28.9 & 22.8 & \gc37.4 & 26.0 & \gc\textbf{71.0} & \textbf{50.0} & \gc62.0 & 47.9\\
            emblem3 & \gc35.8 & 29.0 & \gc29.5 & 18.2 & \gc40.0 & 0.6 & \gc\textbf{52.5} & \textbf{42.0}\\
            emblem4 & \gc2.3  & 1.0  & \gc1.5 & 1.1 & \gc1.7 & 2.2 & \gc\textbf{21.2} & \textbf{20.0}\\
            emblem5 & \gc21.3 & 16.4 & \gc15.1 & 11.2 & \gc\textbf{28.4} & 1.3 & \gc26.1 & \textbf{23.6}\\
            flower1 & \gc0.7  & 0.9  & \gc1.7 & 1.5 & \gc0.9 & 0.0 & \gc\textbf{15.0} & \textbf{13.6}\\
            flower2 & \gc41.2 & 21.1 & \gc36.5 & 19.0 & \gc\gc25.0 & 0.8 & \gc\textbf{50.0} & \textbf{45.9}\\
            flower3 & \gc4.7  & 5.2  & \gc7.7 & 7.5 & \gc\textbf{26.2} & 0.0 & \gc21.8 & \textbf{19.3}\\
            flower4 & \gc2.8  & 2.4  & \gc1.3 & 1.3 & \gc24.1 & 2.1 & \gc\textbf{28.1} & \textbf{27.9}\\
            flower5 & \gc28.6 & 5.4  & \gc18.1 & 5.7 & \gc\textbf{43.4} & \textbf{27.3} & \gc28.9 & 21.1\\
            \midrule
            \vspace{2mm}
        \end{tabular}
        }
    \end{subtable}
    \hfill
    \begin{subtable}[b]{0.513\textwidth}
        \renewcommand{\arraystretch}{1}
        \resizebox{1\linewidth}{!}{
        \begin{tabular}{lcccccccc}
            \toprule
            Method & \multicolumn{2}{c}{Naive} & \multicolumn{2}{c}{VAE~\cite{kingma2013auto}} & \multicolumn{2}{c}{STAE~\cite{chaki2023STAE}} & \multicolumn{2}{c}{Cong.~\cite{cox2008LScongealing}}\\
            \midrule
            Category & \gc C & CU & \gc C & CU & \gc C & CU & \gc C & CU\\
            \midrule
            interlacing1 & \gc1.2 & 0.8  & \gc1.1 & 0.6 & \gc11.9 & 0.0 & \gc\textbf{45.7} & \textbf{12.6}\\
            interlacing2 & \gc\textbf{39.8} & \textbf{27.1} & \gc30.5 & 20.4 & \gc27.6 & 15.9 & \gc36.0 & 26.1 \\
            interlacing3 & \gc31.3 & 25.0 & \gc\textbf{40.2} & \textbf{31.7} & \gc21.9 & 0.0 & \gc22.8 & 14.6 \\
            interlacing4 & \gc32.4 & 17.6 & \gc39.7 & 18.1 & \gc35.5 & 16.0 & \gc\textbf{48.0} & \textbf{23.5} \\
            interlacing5 & \gc2.8 & 1.3  & \gc3.1  & 1.8 & \gc15.6  & 0.3 & \gc\textbf{24.7} & \textbf{7.3}\\
            ring1        & \gc18.5 & 14.8 & \gc17.7 & 10.3 & \gc8.3 & 4.3 & \gc\textbf{37.9} & \textbf{29.8} \\
            ring2        & \gc13.5 & 11.1 & \gc15.9 & 11.8 & \gc\textbf{27.2} & \textbf{19.1} & \gc23.3 & 17.5\\
            ring3        & \gc0.6 & 0.6 & \gc1.4 & 1.2 & \gc1.9 & 1.5 & \gc\textbf{22.9} & \textbf{13.4}\\
            ring4        & \gc1.5 & 1.5 & \gc6.3 & 4.5 & \gc3.9 & 3.8 & \gc\textbf{10.1} & \textbf{7.5} \\
            ring5        & \gc0.5  & 0.8 & \gc1.4 & 1.5 & \gc0.0 & 0.0 & \gc\textbf{5.4} & \textbf{5.1} \\
            symbol1      & \gc0.0 & 0.0 & \gc3.3 & 1.9 & \gc4.9 & 0.0 & \gc\textbf{7.3} &\textbf{5.4} \\
            symbol2      & \gc8.4 & 3.4 & \gc13.6 & 2.9 & \gc7.4 & 6.8& \gc\textbf{16.7} & \textbf{11.9}\\
            symbol3      & \gc2.2 & 1.2 & \gc13.0 & 10.6& \gc29.9 & 14.6 & \gc\textbf{42.7} & \textbf{19.0} \\
            symbol4      & \gc11.9 & 11.3& \gc8.3  &6.9& \gc\textbf{19.0} & 5.1& \gc16.2 & \textbf{14.9}\\
            symbol5      & \gc0.8 & 0.5 & \gc5.1 & 3.3 & \gc\textbf{22.9} & \textbf{23.5} & \gc20.3 & 16.7\\
            \midrule
            mIoU    & \gc13.9 & 9.3 & \gc14.2 & 9.1 & \gc19.1 & 8.0 & \gc\textbf{27.4} & \textbf{20.2}\\
          \bottomrule
            \arrayrulecolor{white}  
            \bottomrule
        \end{tabular}
        }
    \end{subtable}
    \vspace{0.8em}
    \caption{\textbf{Quantitative results for change localization.} We detail the performance of the different baselines on all of the 30 vignette categories. We also report the average value of the IoU (mIoU) on all vignettes.}
    \vspace{-1em}
    \label{tab:change_per_cat_results}
\end{table}

\begin{figure}[t]
    \centering
    \resizebox{\linewidth}{!}{
    \begin{tabular}{ccccccc}
        \changedettable{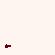} \\
        \changedettable{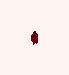} \\
        \changedettable{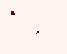} \\
        \changedettable{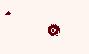} \\
        \changedettable{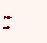} \\
        \vspace{-1.8em}
        \changedettablerot{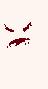} \\
        (a) Normal & (b) Changed & (c) GT & (d) Naive & (e) VAE & (f) STAE & (g) Cong.  \\
    \end{tabular}
    }
    \caption{\textbf{Qualitative results for change localization.} For randomly selected vignettes, we show (a) an example normal vignette as well as (b) the changed vignette with (c) the corresponding ground truth change mask (GT). For each method (d-g), we show the predicted difference image. 
    }
    \vspace{-1em}\label{fig:qualitative_visu_change_det}
\end{figure}

\subsection{Vignettes and unsupervised change localization}

\subsubsection{Methods.}
We evaluate methods that, given a test image, compute its best approximation similar to the reference images, then rely on the difference between the predicted and original image to identify the changes. The most naive approach is to approximate the test image using the average of the reference images, which provides a first baseline. A natural improvement over this method is to perform congealing~\cite{cox2008LScongealing} first on the reference images to obtain an `aligned average', then align it to the test sample before computing the difference. Alignment is done using a color and an affine transformation. We refer to this approach as `congealing'. 

\begin{wrapfigure}{r}{0.5\linewidth}
    \centering
    \vspace{-3.2em}
    \includegraphics[width=\linewidth]{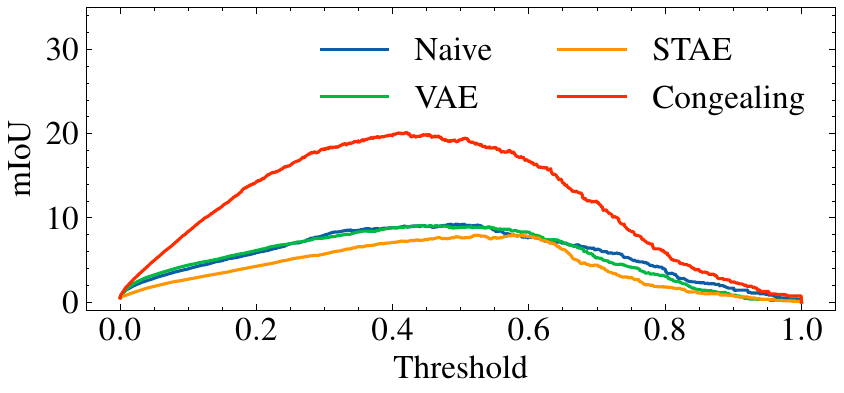}
    \caption{\textbf{Threshold value selection.} We show the mIoU of all change localization methods as a function of the threshold value used to define the changed pixels.}
    \vspace{-3em}
    \label{fig:miou_threshold}
\end{wrapfigure}

We also test more advanced approaches, that rely on learning an auto-encoder on the reference samples and are thus in theory able to learn richer variations, such as the ones related to inking. We evaluate a simple variational auto-encoder (VAE)~\cite{kingma2013auto} using the average of reconstructed images as well as STAE~\cite{chaki2023STAE}, a method combining a spatial transformer (ST) and a fully-connected auto-encoder (AE).
Note that since all these methods compute image differences to localize the changes, computing a segmentation of the changed regions essentially requires deciding on a threshold above which we consider pixels to be changed. We computed the mean IoU (mIoU) for each method for different threshold values, as shown in Figure~\ref{fig:miou_threshold}. We found mIoU to be quite stable for the different threshold values, and thus simply selected the best threshold value for each method. 

\subsubsection{Results.} We report the performance of the different methods in Table~\ref{tab:change_per_cat_results} and show qualitative comparisons in Figure~\ref{fig:qualitative_visu_change_det}. On average, congealing leads to the best results while the other three methods have similar performances. However, looking at the performance of each vignette, the qualitative results paint a slightly different picture. Indeed, STAE performs best in some cases, while having almost zero performance in some others. These performance irregularities seem related to the need for class-dependent threshold values in reconstruction-based methods \cite{venkataramanan2020attention}. 

The qualitative results also hint that morphological operations on the difference images or segmentation maps potentially joined with operations on the original images could improve the quantitative results. However, this would introduce hyper-parameters, that are unlikely to generalize beyond a specific dataset.

\section{Conclusion}
\label{sec:conclusion}
We have introduced a rich and historically meaningful dataset of book ornaments, with annotations and metrics for three unsupervised tasks that are of interest to book historians. We found that despite the apparent simplicity of the printed patterns, complex deep learning methods currently fail to provide satisfying results and in many cases were outperformed by simple approaches. We thus hope this work will have a significant impact both by stimulating the applications of computer vision methods to printed material and by changing the design and evaluation of clustering, element discovery and change localization methods.
\section{Acknowledgement}
This work was funded by ANR ROIi project ANR-20-CE38-0005. S. Baltaci, E. Vincent, and M. Aubry were supported by ERC project DISCOVER funded by the European Union’s Horizon Europe Research and Innovation program under grant agreement No. 101076028 and ANR VHS project ANR-21-CE38-0008. We thank Silya Ounoughi, Thomas Gautrais, and Vincent Ventresque for their work in the collection and annotation of the datasets, and Ségolène Albouy, Raphaël Baena, Syrine Kalleli, Ioannis Siglidis, Gurjeet Sangra Singh, Andrea Morales Garzón and Malamatenia Vlachou for valuable feedbacks. 

{
\bibliographystyle{splncs04}
\bibliography{references.bib}

\begin{thebibliography}{10}
\providecommand{\url}[1]{\texttt{#1}}
\providecommand{\urlprefix}{URL }
\providecommand{\doi}[1]{https://doi.org/#1}

\bibitem{akcay2019ganomaly}
Akcay, S., Atapour-Abarghouei, A., Breckon, T.P.: Ganomaly: Semi-supervised
  anomaly detection via adversarial training. In: Computer Vision--ACCV 2018:
  14th Asian Conference on Computer Vision, Perth, Australia, December 2--6,
  2018, Revised Selected Papers, Part III 14. pp. 622--637. Springer (2019)

\bibitem{BahierPorte2022}
Bahier-Porte, C., Fournel, T., Vial-Bonacci, L., Denis~Emonet, R., Habrard, A.,
  Ventresque, V., Gautrais, T.: Regions of interest to investigate after
  learning the use of ornaments by marc-michel rey. In: Conference Abstracts of
  Digital Humanities 2022, Pannel Computer Vision for the Study of Printers’
  Ornaments and Illustrations in European Hand-Press Books. pp. 66--67 (2022)

\bibitem{BahierPorte2021}
Bahier-Porte, C.: « l'expérience me l'apprend » : Marc michel rey et la
  presse en hollande. In: University of Toronto Quarterly. vol. 89 (4), pp.
  731--746 (2021)

\bibitem{Baudrier2009}
Baudrier, E., Busson, S., Corsini, S., Delalandre, M., Landré, J., ,
  Morain-Nicolier, F.: Retrieval of the ornaments from the hand-press period:
  an overview. In: 10th International Conference on Document Analysis and
  Recognition. IEEE. pp. 496--500 (2009)

\bibitem{baur2019deep}
Baur, C., Wiestler, B., Albarqouni, S., Navab, N.: Deep autoencoding models for
  unsupervised anomaly segmentation in brain mr images. In: Brainlesion:
  Glioma, Multiple Sclerosis, Stroke and Traumatic Brain Injuries: 4th
  International Workshop, BrainLes 2018, Held in Conjunction with MICCAI 2018,
  Granada, Spain, September 16, 2018, Revised Selected Papers, Part I 4. pp.
  161--169. Springer (2019)

\bibitem{bergel2013content}
Bergel, G., Franklin, A., Heaney, M., Arandjelovic, R., Zisserman, A., Funke,
  D.: Content-based image recognition on printed broadside ballads: The
  bodleian libraries' imagematch tool. In: Proceedings of the IFLA World
  Library and Information Congress (2013)

\bibitem{bergmann2019mvtec}
Bergmann, P., Fauser, M., Sattlegger, D., Steger, C.: Mvtec ad--a comprehensive
  real-world dataset for unsupervised anomaly detection. In: Proceedings of the
  IEEE/CVF conference on computer vision and pattern recognition. pp.
  9592--9600 (2019)

\bibitem{Bigun1996}
Bigün, J., Bhattacharjee, S., Michel, S.: Orientation radiograms for image
  retrieval: an alternative to segmentation. In: Proceedings of the ICPR. pp.
  346--350 (1996)

\bibitem{BodleianBallads}
Bodleian ballads: database of woodcuts,
  \url{http://balladsblog.bodleian.ox.ac.uk/blog/1069}

\bibitem{burgess_monet_2019}
Burgess, C.P., Matthey, L., Watters, N., Kabra, R., Higgins, I., Botvinick, M.,
  Lerchner, A.: {{MONet}}: {{Unsupervised Scene Decomposition}} and
  {{Representation}}. arXiv:1901.11390 [cs, stat]  (Jan 2019)

\bibitem{caoSpatiallyCoherentLatent2007}
Cao, L., {Fei-Fei}, L.: Spatially coherent latent topic model for concurrent
  object segmentation and classification. In: {{ICCV}} (2007)

\bibitem{caron_deep_2018}
Caron, M., Bojanowski, P., Joulin, A., Douze, M.: Deep {{Clustering}} for
  {{Unsupervised Learning}} of {{Visual Features}}. In: {{arXiv}}:1807.05520
  [Cs] (ECCV 2018)

\bibitem{caron2021emerging}
Caron, M., Touvron, H., Misra, I., J{\'e}gou, H., Mairal, J., Bojanowski, P.,
  Joulin, A.: Emerging properties in self-supervised vision transformers. In:
  Proceedings of the IEEE/CVF international conference on computer vision. pp.
  9650--9660 (2021)

\bibitem{chaki2023STAE}
Chaki, S., Steinlin, S., Emonet, R., Fournel, T.: One-to-many pattern
  comparison combining fully-connected autoencoder with spatial transformer for
  ornament investigation. https://doi.org/10.21203/rs.3.rs-3573134/v1  (2023)

\bibitem{choUnsupervisedObjectDiscovery2015}
Cho, M., Kwak, S., Schmid, C., Ponce, J.: Unsupervised {{Object Discovery}} and
  {{Localization}} in the {{Wild}}. In: {{CVPR}} (2015)

\bibitem{Chung2014}
Chung, J.S., Arandjelovic, R., Bergel, G., Franklin, A., Zisserman, A.:
  Re-presentations of art collections. In: Workshop on Computer Vision for Art
  Analysis (Visart), ECCV (2014)

\bibitem{Compositor}
Compositor: database of ornaments, \url{https://compositor.bham.ac.uk}

\bibitem{corsini1999}
Corsini, S.: La preuve par les fleurons: analyse comparée du matériel
  ornemental des imprimeurs suisses romands, 1775-1785. In: Centre
  international d'étude du XVIIIe siècle (1999)

\bibitem{Corsini2001}
Corsini, S.: « passe-partout~: banque internationale d’ornements
  d’imprimerie. Bulletin des bibliothèques de France  \textbf{5}, ~73 (2001)

\bibitem{Corsini1988}
Corsini, S.: Vers un corpus des ornements typographiques lausannois du xviiie
  siècle : problèmes de définition et de méthode. In: Ornementation
  typographique et bibliographie historique. vol. Mons et Bruxelles, Van
  Balberghe, pp. 139--158 (1988)

\bibitem{cox2008LScongealing}
Cox, M., Sridharan, S., Lucey, S., Cohn, J.: Least squares congealing for
  unsupervised alignment of images. In: 2008 IEEE Conference on Computer Vision
  and Pattern Recognition. pp.~1--8 (2008). \doi{10.1109/CVPR.2008.4587573}

\bibitem{crawford2019spatially}
Crawford, E., Pineau, J.: Spatially invariant unsupervised object detection
  with convolutional neural networks. In: Proceedings of the AAAI Conference on
  Artificial Intelligence. vol.~33, pp. 3412--3420 (2019)

\bibitem{Dutta2021}
Dutta, A., Bergel, G., Zisserman, A.: Visual analysis of chapbooks printed in
  scotland. In: The 6th International Workshop on Historical Document Imaging
  and Processing. pp. 67--72 (2021)

\bibitem{engelcke_genesis_2020}
Engelcke, M., Kosiorek, A.R., Jones, O.P., Posner, I.: {{GENESIS}}:
  {{Generative Scene Inference}} and {{Sampling}} with {{Object-Centric Latent
  Representations}}. ICLR  (2020)

\bibitem{engelcke2021genesis}
Engelcke, M., Parker~Jones, O., Posner, I.: Genesis-v2: Inferring unordered
  object representations without iterative refinement. Advances in Neural
  Information Processing Systems  \textbf{34},  8085--8094 (2021)

\bibitem{Enschede}
Enschedé, J.: Proef van letteren, welke gegooten worden in de nieuwe
  haerlemsche lettergietery van j. enschedé (1768),
  \url{https://gallica.bnf.fr/ark:/12148/bpt6k328783b}

\bibitem{eslami2016attend}
Eslami, S., Heess, N., Weber, T., Tassa, Y., Szepesvari, D., Hinton, G.E.,
  et~al.: Attend, infer, repeat: Fast scene understanding with generative
  models. Advances in neural information processing systems  \textbf{29} (2016)

\bibitem{pascal}
Everingham, M., Gool, L.V., Williams, C.K.I., Winn, J.M., Zisserman, A.: The
  pascal visual object classes (voc) challenge. Int. J. Comput. Vis.
  \textbf{88}(2),  303--338 (2010),
  \url{http://dblp.uni-trier.de/db/journals/ijcv/ijcv88.html}

\bibitem{mlch2020}
Fiorucci, M., Khoroshiltseva, M., Pontil, M., Traviglia, A., {Del Bue}, A.,
  James, S.: Machine learning for cultural heritage: A survey. Pattern
  Recognition Letters  \textbf{133},  102--108 (2020).
  \doi{https://doi.org/10.1016/j.patrec.2020.02.017},
  \url{https://www.sciencedirect.com/science/article/pii/S0167865520300532}

\bibitem{Fleuron}
Fleuron: database of printing ornaments,
  \url{https://db-prod-bcul.unil.ch/ornements/scripts/index.html}

\bibitem{Fournier}
Fournier, P.S.: Les caractères de l'imprimerie (1764),
  \url{https://gallica.bnf.fr/ark:/12148/bpt6k15021752/}

\bibitem{frey2003tiem}
Frey, B., Jojic, N.: Transformation-invariant clustering using the em
  algorithm. Pattern Analysis and Machine Intelligence, IEEE Transactions on
  \textbf{25},  1-- 17 (02 2003). \doi{10.1109/TPAMI.2003.1159942}

\bibitem{frey2001flti}
Frey, B.J., Jojic, N.: Fast, large-scale transformation-invariant clustering.
  In: Dietterich, T., Becker, S., Ghahramani, Z. (eds.) Advances in Neural
  Information Processing Systems. vol.~14. MIT Press (2001),
  \url{https://proceedings.neurips.cc/paper_files/paper/2001/file/95f6870ff3dcd442254e334a9033d349-Paper.pdf}

\bibitem{Goyal2020APG}
Goyal, K., Dyer, C., Warren, C.N., G'Sell, M.G., Berg-Kirkpatrick, T.: A
  probabilistic generative model for typographical analysis of early modern
  printing. In: Annual Meeting of the Association for Computational Linguistics
  (2020), \url{https://api.semanticscholar.org/CorpusID:218486915}

\bibitem{goyette2012changedetection}
Goyette, N., Jodoin, P.M., Porikli, F., Konrad, J., Ishwar, P.:
  Changedetection. net: A new change detection benchmark dataset. In: 2012 IEEE
  computer society conference on computer vision and pattern recognition
  workshops. pp.~1--8. IEEE (2012)

\bibitem{graumanUnsupervisedLearningCategories2006}
Grauman, K., Darrell, T.: Unsupervised {{Learning}} of {{Categories}} from
  {{Sets}} of {{Partially Matching Image Features}}. In: {{CVPR}} (2006)

\bibitem{greffMultiObjectRepresentationLearning2019}
Greff, K., Kaufman, R.L., Kabra, R., Watters, N., Burgess, C., Zoran, D.,
  Matthey, L., Botvinick, M., Lerchner, A.: Multi-{{Object Representation
  Learning}} with {{Iterative Variational Inference}}. In: {{ICML}} (2019)

\bibitem{ijcai2017p243}
Guo, X., Gao, L., Liu, X., Yin, J.: Improved deep embedded clustering with
  local structure preservation. In: Proceedings of the Twenty-Sixth
  International Joint Conference on Artificial Intelligence, {IJCAI-17}. pp.
  1753--1759 (2017). \doi{10.24963/ijcai.2017/243}

\bibitem{He2016AMG}
He, S., Samara, P., Burgers, J., Schomaker, L.: A multiple-label guided
  clustering algorithm for historical document dating and localization. IEEE
  Transactions on Image Processing  \textbf{25},  5252--5265 (2016),
  \url{https://api.semanticscholar.org/CorpusID:16772542}

\bibitem{hsu2017cnn}
Hsu, C.C., Lin, C.W.: {CNN}-based joint clustering and representation learning
  with feature drift compensation for large-scale image data. IEEE Transactions
  on Multimedia  \textbf{20}(2),  421--429 (2017)

\bibitem{hu2017learning}
Hu, W., Miyato, T., Tokui, S., Matsumoto, E., Sugiyama, M.: Learning discrete
  representations via information maximizing self-augmented training. In:
  International conference on machine learning. pp. 1558--1567. PMLR (2017)

\bibitem{Huang2022DeepCluEEI}
Huang, D., Chen, D., Chen, X., Wang, C., Lai, J.: Deepclue: Enhanced image
  clustering via multi-layer ensembles in deep neural networks. ArXiv
  \textbf{abs/2206.00359} (2022),
  \url{https://api.semanticscholar.org/CorpusID:249240539}

\bibitem{jaderberg_spatial_2015}
Jaderberg, M., Simonyan, K., Zisserman, A.: Spatial {{Transformer Networks}}.
  In: Advances in {{Neural Information Processing Systems}}. vol.~28 (2015)

\bibitem{ji2019invariant}
Ji, X., Henriques, J.F., Vedaldi, A.: Invariant information clustering for
  unsupervised image classification and segmentation. In: Proceedings of the
  IEEE International Conference on Computer Vision. pp. 9865--9874 (2019)

\bibitem{jiang2020generative}
Jiang, J., Ahn, S.: Generative neurosymbolic machines. Advances in Neural
  Information Processing Systems  \textbf{33},  12572--12582 (2020)

\bibitem{johnsonCLEVRDiagnosticDataset2017}
Johnson, J., Hariharan, B., {van der Maaten}, L., {Fei-Fei}, L., Zitnick, C.L.,
  Girshick, R.: {{CLEVR}}: {{A}} diagnostic dataset for compositional language
  and elementary visual reasoning. In: {{CVPR}} (2017)

\bibitem{joulinDiscriminativeClusteringImage2010}
Joulin, A., Bach, F., Ponce, J.: Discriminative clustering for image
  co-segmentation. In: {{CVPR}} (2010)

\bibitem{multiobjectdatasets19}
Kabra, R., Burgess, C., Matthey, L., Kaufman, R.L., Greff, K., Reynolds, M.,
  Lerchner, A.: Multi-object datasets.
  https://github.com/deepmind/multi\_object\_datasets/ (2019)

\bibitem{karazija2021clevrtex}
Karazija, L., Laina, I., Rupprecht, C.: Clevrtex: A texture-rich benchmark for
  unsupervised multi-object segmentation. In: Thirty-fifth Conference on Neural
  Information Processing Systems Datasets and Benchmarks Track (Round 2) (2021)

\bibitem{kingma2013auto}
Kingma, D.P., Welling, M.: Auto-encoding variational bayes. In: 2nd
  International Conference on Learning Representations, {ICLR} 2014, Banff, AB,
  Canada, April 14-16, 2014, Conference Track Proceedings (2014)

\bibitem{kosiorek_stacked_2019}
Kosiorek, A.R., Sabour, S., Teh, Y.W., Hinton, G.E.: Stacked {{Capsule
  Autoencoders}}. In: Advances in {{Neural Information Processing Systems}}.
  vol.~23 (Dec 2019)

\bibitem{Kuhn1955TheHM}
Kuhn, H.W.: The hungarian method for the assignment problem. Naval Research
  Logistics (NRL)  \textbf{52} (1955),
  \url{https://api.semanticscholar.org/CorpusID:9426884}

\bibitem{learned2005data}
Learned-Miller, E.G.: Data driven image models through continuous joint
  alignment. IEEE Transactions on Pattern Analysis and Machine Intelligence
  \textbf{28}(2),  236--250 (2005)

\bibitem{lesjak2018novel}
Lesjak, {\v{Z}}., Galimzianova, A., Koren, A., Lukin, M., Pernu{\v{s}}, F.,
  Likar, B., {\v{S}}piclin, {\v{Z}}.: A novel public mr image dataset of
  multiple sclerosis patients with lesion segmentations based on multi-rater
  consensus. Neuroinformatics  \textbf{16},  51--63 (2018)

\bibitem{Lin2020SPACE}
Lin, Z., Wu, Y.F., Peri, S.V., Sun, W., Singh, G., Deng, F., Jiang, J., Ahn,
  S.: Space: Unsupervised object-oriented scene representation via spatial
  attention and decomposition. In: International Conference on Learning
  Representations (2020), \url{https://openreview.net/forum?id=rkl03ySYDH}

\bibitem{liu2020towards}
Liu, W., Li, R., Zheng, M., Karanam, S., Wu, Z., Bhanu, B., Radke, R.J., Camps,
  O.: Towards visually explaining variational autoencoders. In: Proceedings of
  the IEEE/CVF Conference on Computer Vision and Pattern Recognition. pp.
  8642--8651 (2020)

\bibitem{macqueen1967some}
MacQueen, J., et~al.: Some methods for classification and analysis of
  multivariate observations. In: Proceedings of the fifth Berkeley symposium on
  mathematical statistics and probability. vol.~1, pp. 281--297. Oakland, CA,
  USA (1967)

\bibitem{Maguelone}
Maguelone: database of typographical ornaments,
  \url{http://maguelone.enssib.fr}

\bibitem{McKennaMori2019}
McKenna, A., Mori, G.: Claude-françois simon, imprimeur, dit « poppy »,
  contrefacteur de marc michel rey. In: https://mmrey.hypotheses.org/778 (2019)

\bibitem{metaxas2023divclust}
Metaxas, I.M., Tzimiropoulos, G., Patras, I.: Divclust: Controlling diversity
  in deep clustering. In: Proceedings of the IEEE/CVF Conference on Computer
  Vision and Pattern Recognition. pp. 3418--3428 (2023)

\bibitem{monnier_deep_2020}
Monnier, T., Groueix, T., Aubry, M.: Deep {{Transformation-Invariant
  Clustering}}. In: {{NeurIPS}} (Oct 2020)

\bibitem{monnier_unsupervised_2021}
Monnier, T., Vincent, E., Ponce, J., Aubry, M.: Unsupervised {{Layered Image
  Decomposition}} into {{Object Prototypes}}. In: Proceedings of the
  {{IEEE}}/{{CVF International Conference}} on {{Computer Vision}}. pp.
  8640--8650 (Apr 2021)

\bibitem{naumann2024tampar}
Naumann, A., Hertlein, F., D{\"o}rr, L., Furmans, K.: Tampar: Visual tampering
  detection for parcel logistics in postal supply chains. In: Proceedings of
  the IEEE/CVF Winter Conference on Applications of Computer Vision. pp.
  8076--8086 (2024)

\bibitem{pang2021deep}
Pang, G., Shen, C., Cao, L., Hengel, A.V.D.: Deep learning for anomaly
  detection: A review. ACM computing surveys (CSUR)  \textbf{54}(2),  1--38
  (2021)

\bibitem{pimentel2014review}
Pimentel, M.A., Clifton, D.A., Clifton, L., Tarassenko, L.: A review of novelty
  detection. Signal processing  \textbf{99},  215--249 (2014)

\bibitem{radford2021learning}
Radford, A., Kim, J.W., Hallacy, C., Ramesh, A., Goh, G., Agarwal, S., Sastry,
  G., Askell, A., Mishkin, P., Clark, J., et~al.: Learning transferable visual
  models from natural language supervision. In: International conference on
  machine learning. pp. 8748--8763. PMLR (2021)

\bibitem{Rey_database}
Rey database: database of publishing ornaments, \url{ROIi:
  https://heurist.huma-num.fr/heurist/?db=ROIi}

\bibitem{Riffaud2011}
Riffaud, A., Pantin, I.: Une archéologie du livre français moderne. Droz
  (2011)

\bibitem{Rosart}
Rosart, J.: Epreuves des caractères qui se gravent et se fondent dans la
  nouvelle fonderie de jacques françois rosart (1761)

\bibitem{rubinsteinUnsupervisedJointObject2013}
Rubinstein, M., Joulin, A., Kopf, J., Liu, C.: Unsupervised {{Joint Object
  Discovery}} and {{Segmentation}} in {{Internet Images}}. In: {{CVPR}} (2013)

\bibitem{ruff2021unifying}
Ruff, L., Kauffmann, J.R., Vandermeulen, R.A., Montavon, G., Samek, W., Kloft,
  M., Dietterich, T.G., M{\"u}ller, K.R.: A unifying review of deep and shallow
  anomaly detection. Proceedings of the IEEE  \textbf{109}(5),  756--795 (2021)

\bibitem{russellUsingMultipleSegmentations2006}
Russell, B.C., Freeman, W.T., Efros, A.A., Sivic, J., Zisserman, A.: Using
  {{Multiple Segmentations}} to {{Discover Objects}} and their {{Extent}} in
  {{Image Collections}}. In: {{CVPR}} (2006)

\bibitem{sauvalle2023unsupervised}
Sauvalle, B., de~La~Fortelle, A.: Unsupervised multi-object segmentation using
  attention and soft-argmax. In: Proceedings of the IEEE/CVF Winter Conference
  on Applications of Computer Vision. pp. 3267--3276 (2023)

\bibitem{schlegl2017unsupervised}
Schlegl, T., Seeb{\"o}ck, P., Waldstein, S.M., Schmidt-Erfurth, U., Langs, G.:
  Unsupervised anomaly detection with generative adversarial networks to guide
  marker discovery. In: International conference on information processing in
  medical imaging. pp. 146--157. Springer (2017)

\bibitem{shen2021s2looking}
Shen, L., Lu, Y., Chen, H., Wei, H., Xie, D., Yue, J., Chen, R., Lv, S., Jiang,
  B.: S2looking: A satellite side-looking dataset for building change
  detection. Remote Sensing  \textbf{13}(24), ~5094 (2021)

\bibitem{shen_discovering_2019}
Shen, X., Efros, A.A., Aubry, M.: Discovering {{Visual Patterns}} in {{Art
  Collections}} with {{Spatially-consistent Feature Learning}}. In: Proceedings
  of the {{IEEE Conference}} on {{Computer Vision}} and {{Pattern Recognition}}
  (Mar 2019)

\bibitem{simeoni2021lost}
Sim\'eoni, O., Puy, G., Vo, H.V., Roburin, S., Gidaris, S., Bursuc, A.,
  P\'erez, P., Marlet, R., Ponce, J.: Localizing objects with self-supervised
  transformers and no labels. Proceedings of the British Machine Vision
  Conference (BMVC)  (November 2021)

\bibitem{sivic2008unsupervised}
Sivic, J., Russell, B.C., Zisserman, A., Freeman, W.T., Efros, A.A.:
  Unsupervised discovery of visual object class hierarchies. In: 2008 IEEE
  Conference on Computer Vision and Pattern Recognition. pp.~1--8. IEEE (2008)

\bibitem{smirnov_marionette_2021}
Smirnov, D., Gharbi, M., Fisher, M., Guizilini, V., Efros, A.A., Solomon, J.:
  {{MarioNette}}: {{Self-Supervised Sprite Learning}}. arXiv:2104.14553 [cs]
  (Apr 2021)

\bibitem{vangansbeke2020scan}
Van~Gansbeke, W., Vandenhende, S., Georgoulis, S., Proesmans, M., Van~Gool, L.:
  Scan: Learning to classify images without labels. In: Proceedings of the
  European Conference on Computer Vision (2020)

\bibitem{venkataramanan2020attention}
Venkataramanan, S., Peng, K.C., Singh, R.V., Mahalanobis, A.: Attention guided
  anomaly localization in images. In: European Conference on Computer Vision.
  pp. 485--503. Springer (2020)

\bibitem{voUnsupervisedImageMatching2019}
Vo, H.V., Bach, F., Cho, M., Han, K., LeCun, Y., P{\'e}rez, P., Ponce, J.:
  Unsupervised {{Image Matching}} and {{Object Discovery}} as {{Optimization}}.
  In: {{CVPR}} (2019)

\bibitem{Wilkinson2021}
Wilkinson, H., Briggs, J., Gorissen, D.: Computer vision and the creation of a
  database of printers’ ornaments. Digital Humanities Quarterly  (2021)

\bibitem{Wilkinson2013}
Wilkinson, H.: ‘printers’ flowers as evidence in the identification of
  unknown printers: Two examples from 1715. In: The Library, 7th ser. vol.~14,
  pp. 70--79 (2013)

\bibitem{xie2016unsupervised}
Xie, J., Girshick, R., Farhadi, A.: Unsupervised deep embedding for clustering
  analysis. In: International conference on machine learning. pp. 478--487.
  PMLR (2016)

\bibitem{yang2017towards}
Yang, B., Fu, X., Sidiropoulos, N.D., Hong, M.: Towards k-means-friendly
  spaces: Simultaneous deep learning and clustering. In: international
  conference on machine learning. pp. 3861--3870. PMLR (2017)

\bibitem{yang2022promising}
Yang, Y., Yang, B.: Promising or elusive? unsupervised object segmentation from
  real-world single images. Advances in Neural Information Processing Systems
  \textbf{35},  4722--4735 (2022)

\bibitem{zenati2018efficient}
Zenati, H., Foo, C.S., Lecouat, B., Manek, G., Chandrasekhar, V.R.: Efficient
  gan-based anomaly detection. arXiv preprint arXiv:1802.06222  (2018)

\bibitem{zhu2021gmair}
Zhu, W., Shen, Y., Yu, L., Aguirre~Sanchez, L.P.: Gmair: Unsupervised object
  detection based on spatial attention and gaussian mixture. arXiv e-prints pp.
  arXiv--2106 (2021)

\end{thebibliography}
}
\end{document}